\documentclass[sigconf,nonacm,10pt]{acmart}

\usepackage{hyperref}
\usepackage{subcaption}
\usepackage{cleveref}
\usepackage{xspace}
\crefname{figure}{Figure}{Figures}
\usepackage{float}
\usepackage{enumitem}
\usepackage{longtable}
\usepackage{graphicx}
\usepackage{algorithm2e}
\usepackage{balance} 
\usepackage{xcolor}
\usepackage{tabularx}

\let\proof\relax
\let\endproof\relax

\usepackage{amsthm}

\usepackage{booktabs}

\let\proof
\let\endproof

\theoremstyle{definition}
\newtheorem{definition}{Definition}

\newcommand{\hide}[1]{}
\newcommand{\ourapproach}{\emph{MLDebugger}\xspace}

\newcommand{\debuggingdecisiontrees}{\emph{Debugging Decision Tree}\xspace}

\renewcommand{\paragraph}[1]{\vspace{0.1cm}\noindent \textbf{#1}}

\newcommand{\succeed}{\texttt{succeed}\xspace}
\newcommand{\fail}{\texttt{fail}\xspace}

\newcommand{\modelparameter}{property\xspace}
\newcommand{\Modelparameter}{Property\xspace}
\newcommand{\modelparameters}{properties\xspace}

{\begin{list}{$\bullet$}{%
	\setlength{\labelsep}{2pt}\setlength{\leftmargin}{0pt}%
	\setlength{\labelwidth}{0pt}%
	\setlength{\listparindent}{0pt}}}
{\end{list}}

\begin{document}
\title{Debugging Machine Learning Pipelines}

\author{Raoni Louren\c{c}o}
\affiliation{%
  \institution{New York University}
  }
\email{raoni@nyu.edu}

\author{Juliana Freire}
\affiliation{%
  \institution{New York University}
}
\email{juliana.freire@nyu.edu}

\author{Dennis Shasha}
\affiliation{%
  \institution{New York University}
  }
\email{shasha@courant.nyu.edu}

\begin{abstract}
  Machine learning tasks entail the use of complex computational
  pipelines to reach quantitative and qualitative conclusions. If some
  of the activities in a pipeline produce erroneous or uninformative
  outputs, the pipeline may fail or produce incorrect
  results. Inferring the root cause of failures and unexpected
  behavior is challenging, usually requiring much human thought, and
  is both time consuming and error prone. We propose a new approach
  that makes use of iteration and provenance to automatically infer
  the root causes and derive succinct explanations of
  failures. Through a detailed experimental evaluation, we assess the
  cost, precision, and recall of our approach compared to the state of
  the art. Our source code and experimental data will be available for reproducibility and enhancement.
\end{abstract}

\ccsdesc[300]{Information systems~Data provenance}

\keywords{}

\acmDOI{10.1145/3329486.3329489}

\maketitle

\section{Introduction} \label{sec:intro}

Machine learning pipelines, like many computational processes, are characterized by interdependent modules, associated parameters, data inputs, and outputs from which conclusions are derived. If one or more modules in a pipeline produce erroneous outputs, the conclusions may be incorrect.

Discovering the root cause of failures in a pipeline is challenging because problems can come from many different sources, including bugs in the code, input data, and improper parameter settings. This problem is compounded when multiple pipelines are composed. For example, when a machine learning pipeline feeds predictions to a second analytics pipeline, errors in the machine learning model can lead to erroneous actions taken based on the analytics results. 
Similar challenges arise in pipeline design, when developers need to understand trade-offs and test the effectiveness of their design decisions, such as the choice of a particular learning method or the selection of a training data set.
To debug pipelines, users currently expend considerable effort reasoning about possibly incorrect or sub-optimal settings and then executing new pipeline instances to test hypotheses. This is  tedious, time-consuming, and error-prone.

\paragraph{\ourapproach.} We propose \ourapproach, a method that automatically identifies one or more minimal causes of failures or unsatisfactory performance in machine learning pipelines.
It does so (i) by using the provenance of previous runs of a pipeline (i.e., information about the  runs and their results), and (ii) by proposing and running carefully selected  configurations consisting of so far untested new combinations of parameter values.

\begin{figure}[t]
    \begin{center}
	    \includegraphics[width=0.98\columnwidth]{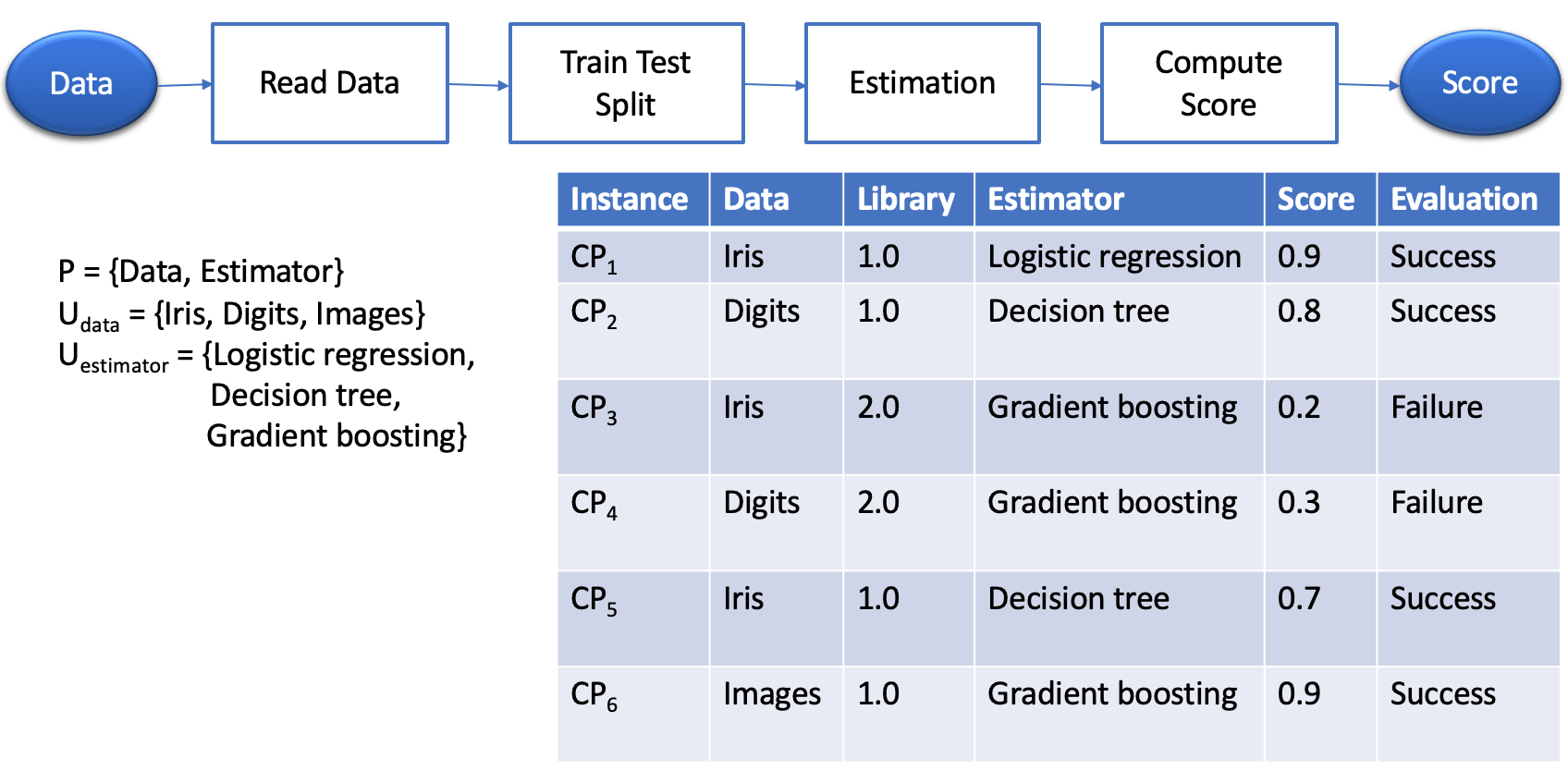}
	\end{center}
	\caption[]{Machine learning pipeline and its provenance. Users can select different input datasets and  classifier estimators to solve a classification problem.}
	\label{fig:pipeline}
\end{figure}

To see why we need to test new configurations, consider a setting in which several analytical
algorithms can be used each with a set of hyperparameters. The results of using some hyperparameter settings can lead to useless outputs (e.g., low quality predictions) or even a crash.  Sometimes, it is unclear which single hyperparameter-value setting or which combinations cause such results.  
Figure~\ref{fig:pipeline} shows a concrete example: a generic template for a machine learning pipeline and a log of different instances of the pipeline that were run with their associated results.
The pipeline reads a data set, splits it into training and test subsets, creates and executes an estimator, and computes the F-measure score using 10-fold cross-validation. 
A data scientist uses this template to understand how different estimators perform for different types of input data, and ultimately, to derive a pipeline instance that leads to high scores. This entails exploring different parameter-values,  training data sets, and  learning classifiers.

Analyzing the provenance of the runs, we can see that
\emph{gradient boosting} leads to much lower scores than other methods for two of the data sets (i.e., Iris and Digits), but it has a high score for the \emph{Images} data. 
This is counter-intuitive and  may suggest that there is a problem with \emph{gradient boosting} for some parameters.
Because each of these runs used different parameters for each method depending on the data set, a definitive conclusion has to await a more systematic exploration through additional experiments with different parameter settings. Furthermore, since many parameter-value combinations can contribute to a poor outcome (or bug), it is also important to derive concise, ideally minimal explanations for the behavior.

Understanding why the values are lower can help identify bugs. In this case,  as we discuss in Section~\ref{sec:strategy}, \ourapproach discovers that the low scores for Gradient Boosting happen only when a specific version the machine learning library containing the estimators is used, suggesting that a bug may have been introduced in this version.

We note that, in the above example, the problem was traced back to a library upgrade. Sometimes, however, the problem can be related to values assigned to variables in specific functions or the data set itself. For example, an industrial colleague cited an example in which an input to an analysis pipeline changed its resolution from monthly to weekly, causing  the analysis to produce erroneous results. Our approach would identify that change in the data to be the root cause of the error. In fact, as  Section~\ref{sec:motivation} shows, our approach 
captures the most relevant aspects of a pipeline, including data, data type, library versions, and values assigned to input parameters.

\paragraph{Contributions}.  
We present a bug-finding and explanation method that makes two novel contributions: 

\begin{enumerate}
    \item 
    Our algorithm (i) takes a given set of pipeline instances some of which give erroneous results, forms a hypothesis about possible root causes, then (ii) carefully selects new pipeline instances 
    to test so-far untested \modelparameters-value combinations. It then iterates on (i) and (ii) until some time budget is exhausted or until it finds a definitive root cause.
    \item 
    From a set of definitive root causes, our method finds and reports a minimal root cause, represented as a Boolean formula containing a minimal set of \modelparameter-comparator-values that would cause the bug.
\end{enumerate}

Because relying solely on stored provenance can lead to incomplete (or incorrect) explanations, Contribution 1 ensures fewer false positives compared to the state of the art. Contribution 2 helps with the precision of the diagnosis, which is necessary for the swift resolution of the bug. 

We also carry out a detailed experimental evaluation which shows that our approach attains higher precision and recall compared with the state of the art, and that it also derives more concise explanations. The experimental data and source code for our system will be made available as open source for reproducibility and enhancement.
We note that the algorithms of \ourapproach are general and can be applied to computational pipelines other than machine learning. Such applications are, however, out of the scope of this paper.

\paragraph{Outline.} The remainder of this paper is organized as follows.  Section~\ref{sec:motivation} introduces the model we use for machine learning pipelines and formally defines the problem we address. In Section~\ref{sec:strategy}, we present algorithms to search for simple and complex causes of failures. We compare our approach with the state of the art in Section~\ref{sec:experiments}.
We review related work in Section~\ref{sec:relatedWork}, and conclude in Section~\ref{sec:conclusion}, where we outline directions for future work.

\section{Definitions and Problem Statement}\label{sec:motivation}

Intuitively, given a set of pipeline instances, some of which have led to bad or questionable results, our goal is to find the root causes of these results possibly by creating and executing new pipeline instances.

\begin{definition}[Pipeline, pipeline instance, \modelparameter-value pairs, value universe, results]
A \textbf{machine learning pipeline} $MP$ is associated with a set of \modelparameters $P$ (i.e., including hyperparameters, input data, versions of programs, computational modules) each of which can take on various values.  A \textbf{pipeline instance}, denoted  as $MP_i$, of $MP$  defines values for the \modelparameters. Thus, an instance $MP_i$ is associated with a list of \textbf{\modelparameter-value pairs} $Pv_i$ containing some assignment $(p,v)$ for all $p \in P$. For each \modelparameter $p \in P$, the \textbf{\modelparameter-value universe} $U_p$ is the set  of all values that have been assigned to $p$ by any pipeline instance so far, i.e., $U_p = \{ v | \exists i (p,v) \in P_i \}$. 
\end{definition}

Note that in a normal use case, our goal is to find the root causes for problematic instances, not to do software verification. Therefore our universe of \modelparameter values for each \modelparameter $p$ is $U_p$, the set already seen. That is,  we seek to understand a root cause from among the \emph{existing} values.

An application-dependent evaluation procedure can be defined to decide whether the pipeline results are acceptable or not, and flag instances that should be investigated. In a machine learning context, an evaluation procedure may be to test whether the cross-validation accuracy is above a certain threshold.

\begin{definition}[Evaluation]
Let $E$ be a procedure that \textbf{evaluates} the results of an instance such that $E(MP_i)=\succeed$ if the results are acceptable, and $E(MP_i)=\fail$ otherwise. 
\end{definition}

\begin{definition}[Hypothetical root cause of failure]
Given a set of instances $G = MP_1, ... , MP_k$ and associated evaluations $E(MP_1 ), .... , E(MP_k )$, a hypothetical root cause of failure is a set 
$C_f$  consisting of a Boolean conjunction of \modelparameter-comparator-value triples which obey the following properties among the instances $G$: (i)  there is at least one $MP_i$ such that  $Pv_i$ satisfies  $C_f$ and  $E(MP_i ) = \fail$; and (ii) if $E(MP_i ) = \succeed$, then  the \modelparameter-values pairs $Pv_i$ of $MP_i$ do not satisfy the conjunction $C_f$.
\end{definition}

\noindent To illustrate the converse of point (ii), if $C_f$ = $A > 5$ and $B =7$, and $MP_i$ has the \modelparameter values $A=15$ and $B=7$ and succeeds, then $C_f$ is unacceptable as a hypothetical root cause of failure.
Given a hypothetical root cause $D$, our framework derives new pipeline instances using different combinations of \modelparameter values to confirm whether $D$ is a definitive root cause.

\begin{definition} [Definitive root cause of failure]
A \emph{hypothetical root cause of failure} $D$ is a 
\emph{definitive root cause of failure} if there is no  instance $MP_q$ from the universe of $U_p$ for each \modelparameter $p$  such that $E(CP_q ) = \succeed$ and $Pv_q$ satisfies $D$. That is, $D$ does not lead to false positives.
\end{definition}

\begin{definition} [Minimal Definitive Root Cause of Failure]
A definitive root cause $D$ is minimal if no proper subset of $D$ is a definitive root cause.
\end{definition}

The example in Figure~\ref{fig:pipeline} illustrates these concepts using the simple machine learning pipeline. 
A possible evaluation procedure would test whether the resulting score is greater than 0.6. In this case, 
\texttt{Data} being different from \emph{Images} and \texttt{Estimator} equal to \emph{gradient boosting} is a hypothetical root cause of failure. Section~\ref{sec:strategy} presents algorithms that determine if this root cause is definitive and minimal.

\paragraph{Problem Definition.} Given a machine learning pipeline $MP$ and a set of \modelparameter-value pairs associated with previously run instances of $MP$,  our goal is to derive minimal definitive root causes.

\section{Debugging Strategy Overview}\label{sec:strategy}

Given a set of pipeline instances, \ourapproach derives
minimal root causes of the problematic instances. Since
trying every possible \modelparameter-value pair combination of the \modelparameter-value universe (an approach that is exponential in the number of \modelparameters) is not feasible in practice,  
\ourapproach uses heuristics that are effective at finding promising configurations. In addition, several causes may contribute to a problem, thus the derived explanations must be concise so that users can understand and act on them.

\ourapproach uses an iterative debugging algorithm called \debuggingdecisiontrees, presented in Section~\ref{sec:decision}.  It discovers simple and complex root causes that can involve a single or multiple \modelparameters and possibly inequalities. 
Because the results of the \debuggingdecisiontrees algorithm consist of disjunctions of conjunctions, they may contain redundancies which we simplify using a heuristic approximating the Quine-McCluskey algorithm described in Section~\ref{sec:simplifying}. 

Intuitively, our method works as follows. Given an initial set of instances, some of which lead to bad outcomes, the algorithms generate new \modelparameter-value configurations (from the same universe) for the suspect instances and combine them first with \modelparameter-values that led to good outcomes. 
That approach has the benefit of swiftly eliminating hypothetical minimal root causes that are not confirmed by the newly generated instances.
While instances can be manually derived by users running instances of the workflow, an initial set of experiments can also be generated by 
random combinations of \modelparameter values, or combinatorial design~\cite{JCD:JCD20065}.

\subsection{Debugging Decision Trees}\label{sec:decision}

An {\em instance} consists of a conjunction of \modelparameter-values and an evaluation (success or failure). A \debuggingdecisiontrees is derived by applying a standard decision tree learning algorithm to all such instances. 
Leaves of the decision tree are either (i) purely \emph{true}, if all pipeline instances leading to a leaf evaluate to \succeed, (ii) purely \emph{false}, if all pipeline instances leading to a leaf evaluate to \fail, or (iii) \emph{mixed}.  

The \debuggingdecisiontrees algorithm works as follows:

\begin{enumerate}
\item Given an initial set of instances $MPI$, construct a decision tree based on the evaluation results for those instances (\succeed or \fail). An inner node of the decision tree is a triple (\textit{\Modelparameter},\textit{Comparator},\textit{Value}), where the \textit{Comparator} indicates whether a given \textit{\Modelparameter} has a value equal to, greater than (or equal to), less than (or equal to), or unequal to \textit{Value}.  
\item If a conjunction (a path in the tree) involving a set of \modelparameters, say, $P_1$ $P_2$, and $P_3$ leads to a consistently failing execution (a pure leaf in decision tree terms), then that combination becomes a ``suspect''.
\item Each suspect combination is used as a filter in a Cartesian product of the \modelparameter values from which new instances will be sampled. 
For example, suppose $P_1=v_1$, $P_2=v_2$, and $P_3=v_3$ is a suspect. To test this suspect, all other \modelparameters will be varied. If every instance having the \modelparameter-values $P_1=v_1$, $P_2=v_2$, and $P_3 = v_3$ leads to failure, then  that conjunction constitutes a \emph{definitive root cause of failure}. 
If the suspect conjunction includes non-equality comparators (e.g., $P_1=v_1$, $P_2=v_2$, and $P_3 > 6$), then we can choose any value for \modelparameters that satisfy the inequalities as an example, 
(e.g., $P_3=7$ or $P_3=8$) and choose pipeline instances having those values. 
Conversely, if any of the newly-generated instances presents a good (\succeed) pipeline instance, the decision tree is rebuilt taking into account the whole set of executed pipeline instances $MPI$  and a new suspect path (one leading to a pure fail outcome) is tried.
\end{enumerate}

Note that if the values associated with a \modelparameter are continuous, \ourapproach starts by choosing the values already attempted. Further analysis can sample other values to uncover additional bugs, but, as mentioned above, our purpose here is to understand the problems already uncovered rather than to verify the software which is of course undecidable in general~\cite{Berger1994}.

Below we present a simple example that illustrates how the \debuggingdecisiontrees algorithm works.

\begin{example}\label{exemp:pipeline}

Consider again the machine learning pipeline in Figure~\ref{fig:pipeline}.
Here, the user is interested in investigating pipelines that lead to low F-measure scores and defines an evaluation function that returns \succeed if $\textit{score} \ge 0.6$.  

For this pipeline, the user is interested in investigating three \modelparameters:
\texttt{Dataset}, the input data to be classified; \texttt{Estimator}, the classification algorithm to be executed; and \texttt{Library Version}, \modelparameter that indicates the version of the machine learning library used. Table~\ref{tab:traces} shows examples of three executions of the pipeline. 
\end{example}

\begin{table}
\centering
\caption{Initial set of classification pipelines instances}
\label{tab:traces}
\begin{tabular}{|c|p{1.7cm}|p{1.3cm}|c|p{1.75cm}|}
\hline
\textbf{Dataset} & \textbf{Estimator} & \textbf{Library Version} & \textbf{Score} & \textbf{Evaluation} ($\textit{score} \ge 0.6$)  \\ \hline
Iris&  Logistic Regression&  1.0&  0.9& \succeed            \\
\hline
Digits& Decision Tree&  1.0&  0.8& \succeed            \\
\hline
Iris& Gradient Boosting&  2.0&  0.2& \fail            \\
\hline
\end{tabular}
\end{table}

\begin{table}
\centering
\caption{Set of classification pipelines instances including the new instances created by \debuggingdecisiontrees based on triple (\textit{Estimator},\textit{Equals to},\textit{Gradient Boosting})}
\label{tab:new_traces_estimator}
\begin{tabular}{|c|p{1.7cm}|p{1.3cm}|c|p{1.75cm}|}
\hline
\textbf{Dataset} & \textbf{Estimator} & \textbf{Library Version} & \textbf{Score} & \textbf{Evaluation} ($\textit{score} \ge 0.6$)  \\ \hline
Iris&  Logistic Regression&  1.0&  0.9& \succeed            \\
\hline
Digits& Decision Tree&  1.0&  0.8& \succeed            \\
\hline
Iris& Gradient Boosting&  2.0&  0.2& \fail            \\
\hline
Digits& Gradient Boosting&  2.0&  0.2& \fail            \\
\hline
Digits& Gradient Boosting&  1.0&  0.7& \succeed            \\
\hline
\end{tabular}
\end{table}

\begin{table}
\centering
\caption{Set of classification pipelines instances including the new instances created by \debuggingdecisiontrees based on triple (\textit{Library Version},\textit{Equals to},\textit{2.0})}
\label{tab:new_traces_version}
\begin{tabular}{|c|p{1.7cm}|p{1.3cm}|c|p{1.75cm}|}
\hline
\textbf{Dataset} & \textbf{Estimator} & \textbf{Library Version} & \textbf{Score} & \textbf{Evaluation} ($\textit{score} \ge 0.6$)  \\ \hline
Iris&  Logistic Regression&  1.0&  0.9& \succeed            \\
\hline
Digits& Decision Tree&  1.0&  0.8& \succeed            \\
\hline
Iris& Gradient Boosting&  2.0&  0.2& \fail            \\
\hline
Digits& Gradient Boosting&  2.0&  0.2& \fail            \\
\hline
Digits& Gradient Boosting&  1.0&  0.7& \succeed            \\
\hline
Digits& Logistic Regression&  2.0&  0.3& \fail            \\
\hline
Iris&  Decision Tree&  2.0&  0.1& \fail            \\
\hline
\end{tabular}
\end{table}

A decision tree is created from the instances shown in Table~\ref{tab:traces} 
that contains a single node: (\textit{Estimator},\textit{Equals to},\textit{Gradient Boosting}).
After assembling new configurations where this triple is true (Table~\ref{tab:new_traces_estimator}), the \debuggingdecisiontrees algorithm observes that new instances present mixed results. 
Hence, we eliminate the hypothetical cause for \texttt{Estimator} with value ``Gradient Boosting'' and the decision tree is rebuilt. After rebuilding, the algorithm finds a new single node tree with the triple (\texttt{Library Version}, \textit{Equals to},$2.0$), indicating a potential problem in that version of the library.
Additional instances are then created to inspect the new root cause candidate, they all \fail as can be seen in Table~\ref{tab:new_traces_version}, confirming the hypothesis, which is output as a definitive root cause.    

\subsection{Simplifying Explanations}\label{sec:simplifying}

Decision trees are easy to read, but they do not always provide minimal explanations. For example,  
we may have two paths leading to pure \emph{false} leaves that differ only in the values of the 
first \modelparameter which takes just two values. Such paths can be reduced to a single conjunction consisting of the \modelparameter-values they share. 
To generate concise explanations from the decision tree, we apply the Quine-McCluskey algorithm~\cite{DBLP:journals/corr/Huang14c}, which provides a method to minimize Boolean functions. Because the algorithm is exponential and encodes the Set Cover problem which is NP-complete,  we use heuristics that do not achieve complete minimality but still reduce the size of the explanation. We illustrate this process below in Example~\ref{exemp:reduction} and evaluate the effectiveness of the simplification in Section~\ref{sec:experiments}.
\hide{need to point to the experiments that show the effectiveness of this simplification} 
Because the use of Quine-McCluskey is not our research contribution, our explanation is brief.

\hide{Reviewer 2: Related to the above, please clarify in Example 2, whether the two conjunctions are supposed to be form, together, a disjunction that is the root cause. Or, are these two different root causes?}

\begin{figure}[t]
    \begin{center}
	    \includegraphics[width=.95\columnwidth]{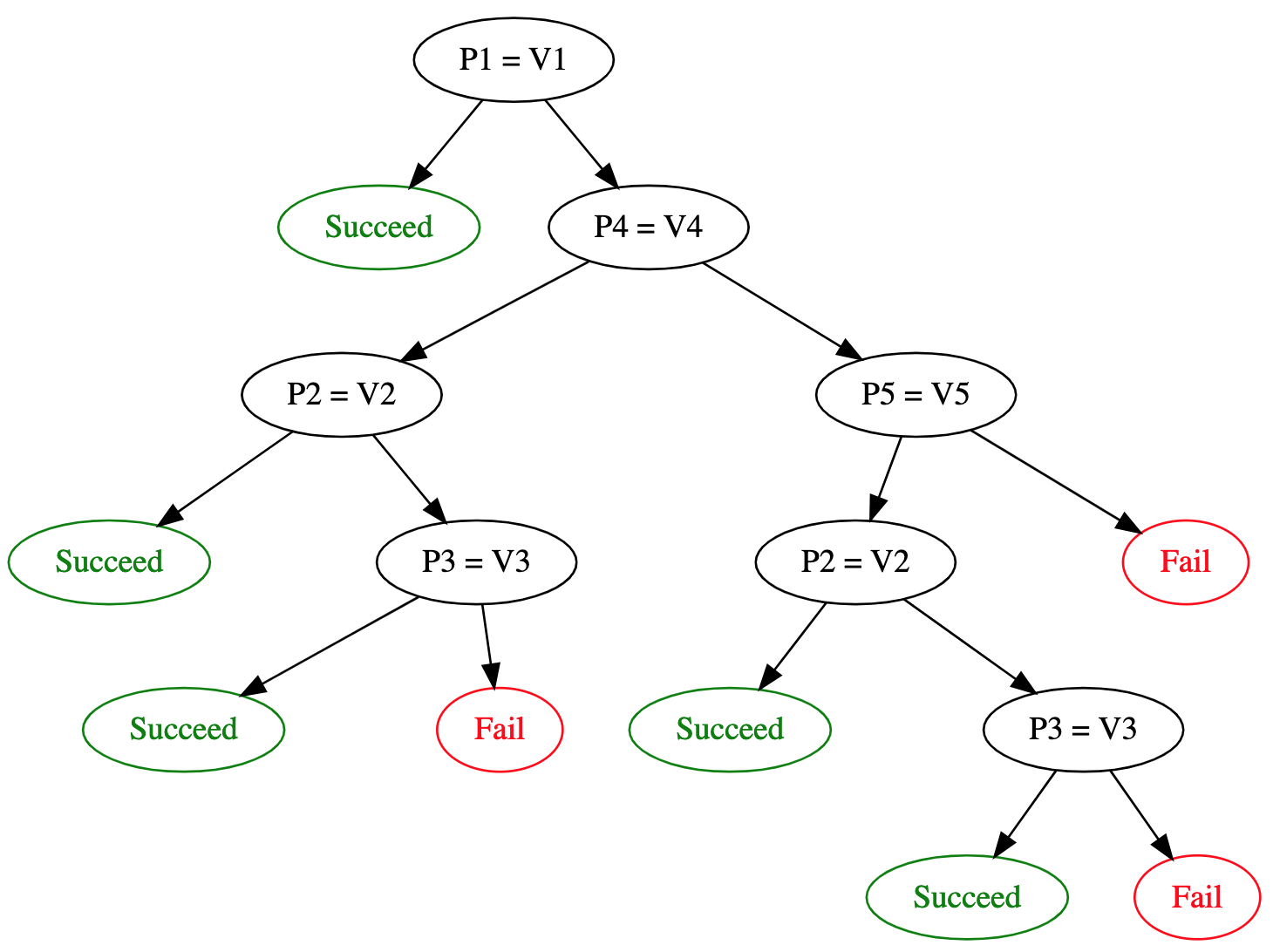}
	\end{center}
	\caption[]{Decision tree fitted over instances. Right branches correspond to equality. Left branches to inequality. Evaluations to \fail occur if ($P1=V1$, $P4 \neq V4 $, $P2=V2$, and $P3=V3$) or ($P1=V_1$, $P4=V4$, and $P5=V5$) or ($P1=V1$, $P4=V4$,  $P5 \neq V5$, $P2=V2$, and $P3=V3$).
	}
	\label{fig:tree}
\end{figure}

\begin{example}\label{exemp:reduction}
Consider an experiment whose instances lead to the decision tree shown in Figure~\ref{fig:tree}.
There are three paths through the tree that evaluate to pure fail outcomes. The Quine-McCluskey algorithm  attempts to shorten these paths to a simpler expression or expressions. The output of the algorithm contains the following disjunction of conjunctions (either one of them will constitute a minimal definitive root cause, but both of which should receive debugging attention):
\[((P1,V1) \land (P2,V2) \land (P3,V3)) \lor 
((P_1,V1) \land (P4,V4) \land (P5,V5))\]

\end{example}

\section{Experimental Evaluation}\label{sec:experiments}

To evaluate the effectiveness of \ourapproach, we compare it against state-of-the-art methods for deriving explanations using a benchmark of machine learning pipeline templates for different tasks.
We examine different scenarios, including when a single minimal definitive root cause is sought (which may be one of several) and when a budget for the number of instances that can be run is set. 

\subsection{Experimental Setup}

\paragraph{Baseline Methods.}
We use two methods for deriving explanations as baselines: Data X-Ray~\cite{Wang:2015:DXD:2723372.2750549} and Explanation Tables~\cite{GebalyFGKS14}. Both analyze the provenance of the pipelines, i.e., the instances previously run and their results, but do not suggest new ones. For that reason, to generate pipeline instances for explanation methods, we gave to each explanation method the instances generated by \ourapproach and by the Sequential Model-Based Algorithm Configuration (SMAC)~\cite{HutHooLey11-smac}. SMAC is an iterative method for hyperparameter optimization that has been shown to be effective compared to previous methods~\cite{Bergstra:2012:RSH:2188385.2188395}. Normally, SMAC looks for good instances, but for debugging purposes, we change its goal to look for bad pipeline instances. 
Note that SMAC proposes new pipeline instances in an iterative fashion, but it always outputs a complete pipeline instance (containing value assignments for all \modelparameters): the best it can find given a budget of instances to run and a criterion. This makes sense for SMAC's primary use case, which is to find a set of parameters that performs well, but it is less helpful for debugging, because a complete pipeline instance is rarely a minimal root cause.
In summary, we combine the explanations with the generative methods: applying Data X-Ray and Explanation Tables to suggest root causes for the pipeline instances generated by SMAC and \ourapproach. 

We also ran experiments using random search as an alternative, i.e., randomly generating instances and then analyzing them. However, the results were always worse than those obtained using SMAC or \ourapproach. Therefore, for simplicity of presentation and to avoid cluttering the plots, we omit these results.  

\begin{figure}[ht]
    \begin{center}
	    \includegraphics[width=.85\columnwidth]{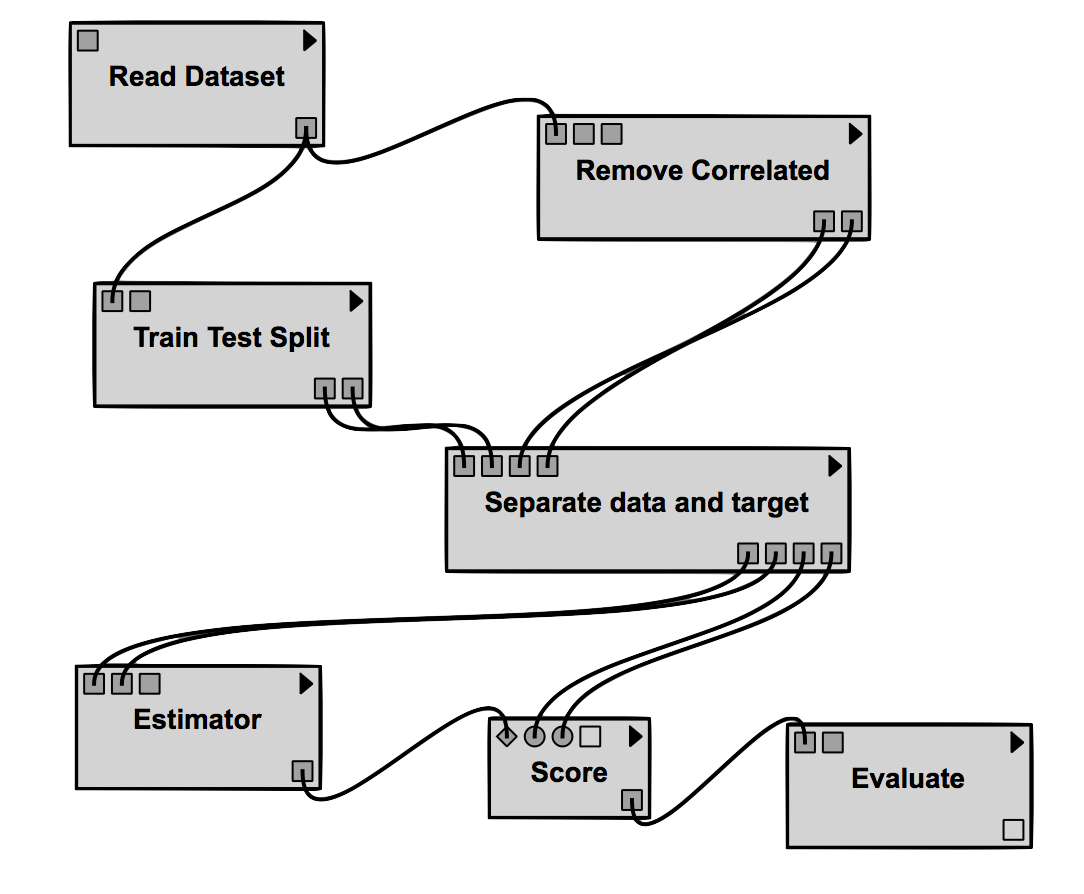}
	\end{center}
	\caption[]{Pipeline template for classification and regression tasks}
	\label{fig:template}
\end{figure}

\paragraph{Machine Learning Pipeline Benchmark.}
We generated a benchmark using the pipeline in Figure~\ref{fig:template}, whose structure is similar to that of Example~\ref{exemp:pipeline}.
It solves the tasks of classification and regression, and it can be used as a template for Kaggle competitions~\cite{Kaggle}. As we describe below, we experimented with three different Kaggle competitions.

We define a threshold for acceptable performance. A pipeline instance that achieves or exceeds that threshold is good. Those that do not are considered to be bad. A Kaggle contestant may use these results to avoid bad parameter values, thus reducing the search space to explore.
For instance, consider the dataset for a Kaggle competition regarding life insurance assessment\cite{KaggleInsurance}, and an accuracy threshold of $0.4$.  \ourapproach identifies two root causes for bad instances (i.e., instances yielding a threshold less than 0.4):

\begin{enumerate}[label=\textbf{I.\arabic*}]
\item \label{r1} $\textit{Estimator} = \text{Gaussian NB}$
\item \label{r2} $\textit{Estimator} = \text{K-Neighbors Classifier}$
\end{enumerate}

\noindent If we increase the value of the accuracy threshold, then \ourapproach finds the following root cause:

\begin{enumerate}[label=\textbf{II.\arabic*}]
\item \label{r3} $\textit{Estimator} \neq \text{Random Forest}$
\end{enumerate}

In addition to the life insurance dataset, we selected: a Kaggle classification competition~\cite{KaggleCancer} whose goal it is to diagnose breast cancer based on a dataset from the University of Wisconsin, and a regression competition~\cite{KaggleRestaurant} whose goal is to predict annual sales revenue of restaurants.  

For each competition, we applied the machine learning pipeline template of Figure~\ref{fig:template}, varying the range of acceptable accuracy score, for classification tasks, and of acceptable R-squared score, for the regression task.

We performed an exhaustive parameter exploration and manually investigated the root causes for failures to create a ground truth for our experiments, so that quality metrics can be computed.

\paragraph{Evaluation Criteria.}
We considered two goals: (i) {\em FindOne} find at least one minimal root cause; (ii) {\em FindAll} find all minimal root causes. The use case for (i) is a debugging setting where it might be useful to work on one bug at a time in the hopes that resolving one may resolve others. The use case for (ii) is when a team of debuggers can work on many bugs in parallel.
 {\em FindAll} may also be useful to provide an overview of the set of issues encountered.

We used the following criteria to measure quality: precision, which measures  the fraction of causes identified by any given method that are in fact minimal definitive root causes; and recall, which measures  (i) in the FindOne case, the fraction of pipelines for which at least one minimal definitive root cause is found and (ii) in the FindAll case, the fraction of all minimal root causes that are found.  
We also report the F-measure, i.e., the harmonic mean of precision and recall:

Formally, let $UMP$ be a set of machine learning pipelines, where each pipeline $MP \in UMP$ (for example the pipeline of Figure~\ref{fig:template}) is associated with a set of minimal definitive root causes $R(MP)$. 
Given a set of minimal root causes asserted by an algorithm $A$,  precision is the number of minimal root causes predicted by  $A$ that are truly minimal definitive root causes (true positives) divided by the size of the set of all root causes asserted by $A$ over all $MP$ in $UMP$. Precision is thus defined as:
\begin{equation*}
   \textit{Precision}=\frac{\sum_{MP \in UMP} |A(MP) \cap R(MP)| }{\sum_{MP \in UMP} |A(MP)|}
\end{equation*}
\noindent where  $A(MP) \cap R(MP) \neq \emptyset$ evaluates to 1 if $A(MP)$ corresponds to at least one of the conjuncts in $R(MP)$.

For the FindOne scenario, recall is the fraction of the $|UMP|$ pipelines when a true minimal definitive root cause is found by $A$:

\begin{equation*}
   \textit{Recall for FindOne}=\frac{\sum_{MP \in UMP} A(MP) \cap R(MP) \neq \emptyset }{|UMP|} 
\end{equation*}
\noindent where  $A(MP) \cap R(MP) \neq \emptyset$ evaluates to 1 if $A(MP)$ corresponds to at least one of the conjuncts in $R(MP)$.

In the FindAll scenario, recall is the fraction of all 
the $R(MP)$ minimal root causes, for all $MP \in UMP$, that are found by the algorithms:

\begin{equation*}
   \textit{Recall for FindAll}=\frac{\sum_{MP \in UMP} |A(MP) \cap R(MP)|}{\sum_{MP \in UMP} |R(MP)|} 
\end{equation*}

\begin{equation*}
   \textit{F-measure}=2\times\frac{\textit{Precision}\times\textit{Recall}}{\textit{Precision}+\textit{Recall}} 
\end{equation*}

Our first set of tests allow \ourapproach to find at least one minimal definitive root cause and then uses the same number of instances for the Data X-Ray and Explanation Tables. Thus, it gives the same budget to each algorithm and checks its precision and recall for the FindOne case. A second set of tests tries different budgets of pipeline instances and evaluates how each algorithm performs in terms of
these same quality metrics.
In these tests, Data X-Ray and Explanation Tables are given (i) the instances  generated by \ourapproach and, in a separate test, (ii) the instances generated by SMAC.
A similar pair of tests is performed for the FindAll case.

\paragraph{Implementation}
The current prototype of \ourapproach is implemented in Python 2.7. It contains a dispatching component which runs in a single thread and spawns multiple pipeline instances in parallel.
In our experiments, we used five execution engine workers to execute the pipeline instances.   

We used the SMAC version for Python 3.6. We also used the   code, implemented by the respective authors, for both the Data X-Ray algorithm (implemented in Java 7)~\cite{Wang:2015:DXD:2723372.2750549} and Explanation Tables~\cite{GebalyFGKS14} (written in python 2.7).
As described above, we used the pipeline instances generated by both \ourapproach and SMAC as inputs to Data X-Ray and Explanation Tables.

The machine learning pipelines were constructed as VisTrails workflows\footnote{\url{www.vistrails.org}}, which allow us to capture the execution provenance of all instances.

All experiments were run on a Linux Desktop (Ubuntu 14.04, 32GB RAM, 3.5GHz $\times$ 8 processor). For purposes of reproducibility and community use, we will make  our code and experiments available.\footnote{\url{https://github.com/raonilourenco/MLDebugger}}

\subsection{Results}
\label{sec:case-study-kaggle}

\begin{figure}[ht]
    \begin{center}
	    \includegraphics[width=.65\columnwidth ]{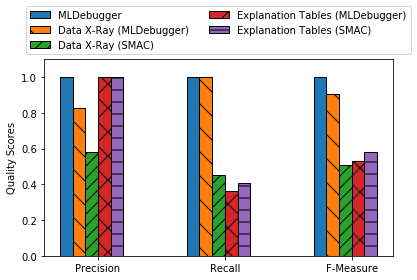}
	\end{center}
	\caption[]{Machine Learning Pipeline Template case study for FindOne. \ourapproach and Explanation Tables have perfect precision, while Data X-Ray returns some false positives.  \ourapproach and Data X-ray using the instances generated by \ourapproach achieve perfect recall, which means that they always find at least one minimal root cause.
	}
	\label{fig:template_plot}
\end{figure}

For our first set of experiments for FindOne, we set \ourapproach to stop iterating as soon as it found one minimal definitive root cause  for failure. 
Figure~\ref{fig:template_plot} shows that \ourapproach and Explanation Tables both achieve perfect precision for the FindOne problem using the instances generated by \ourapproach when that is allowed run until it finds at least one minimal definitive root cause. Data X-Ray finds not only definitive root causes, but also configurations that do not always yield bad instances, resulting in its lower precision. \ourapproach enjoys higher recall, due to its ability to capture also root causes whose comparators are negations or inequalities.

\begin{figure*}[ht]
  \centering
  \subcaptionbox{Precision
  \label{fig:template_budget:a}}{
    \includegraphics[width=0.65\columnwidth]{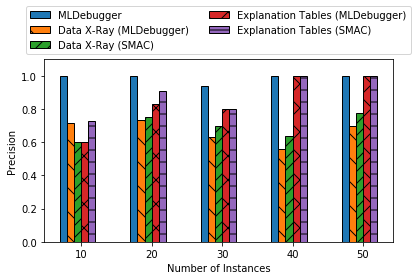}
  }
  \subcaptionbox{Recall 
  \label{fig:template_budget:b}}{
    \includegraphics[width=0.65\columnwidth]{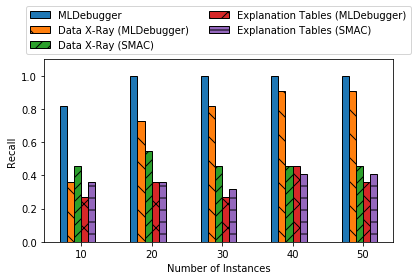}
  }
  \subcaptionbox{F-measure\label{fig:template_budget:c}}{
    \includegraphics[width=0.65\columnwidth]{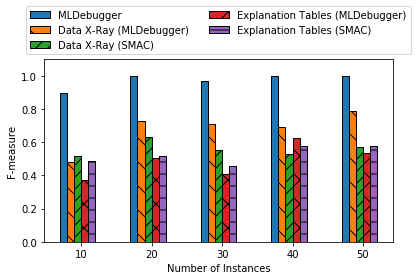}
  }
  \caption{FindOne problem. Machine Learning Pipeline Template case study on a budget. When we  debug the Machine Learning Pipeline on a budget, \ourapproach provides answers with perfect precision, which is also achieved by Explanation Tables with more instances. \ourapproach shows perfect recall, hence f-measure as well, except when running on a very low budget}
  \label{fig:template_budget}
\end{figure*}

Running pipelines can be expensive, so executing a very large number of instances to find a definitive root cause may not be feasible in practice. Hence, we also evaluated the effectiveness of the different methods when a budget is defined for the maximum number of  instances to be executed for the FindOne problem.
Figure~\ref{fig:template_budget} shows the results 
for different budgets for FindOne. Not surprisingly, \ourapproach  makes mistakes  when there is insufficient data to characterize the minimal definitive root causes, being as precise as Explanation Tables on high budget and attaining higher recall throughout.

\begin{figure}[ht]
    \begin{center}
	    \includegraphics[width=0.65\columnwidth ]{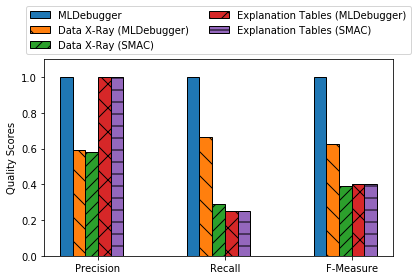}
	\end{center}
	\caption[]{Machine Learning Pipeline Template case study for FindAll. As in the FindOne problem \ourapproach and Explanation Tables have perfect precision, while Data X-Ray returns some false positives. \ourapproach attains higher recall due to its ability to handle inequalities and negations.
	}
	\label{fig:template_plot_all}
\end{figure}

\begin{figure*}[ht]
  \centering
  \subcaptionbox{Precision
  \label{fig:template_budget_all:a}}{
    \includegraphics[width=0.65\columnwidth]{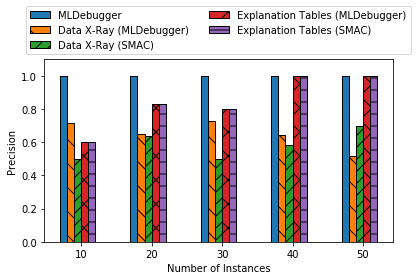}
  }
  \subcaptionbox{Recall 
  \label{fig:template_budget_all:b}}{
    \includegraphics[width=0.65\columnwidth]{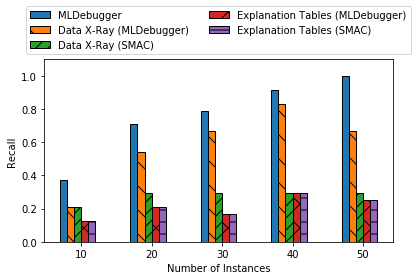}
  }
  \subcaptionbox{F-measure\label{fig:template_budget_all:c}}{
    \includegraphics[width=0.65\columnwidth]{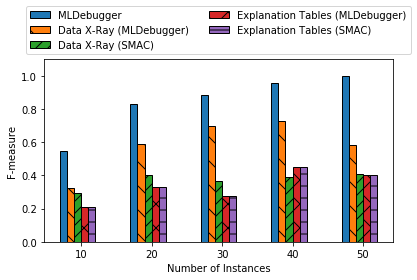}
  }
  \caption{FindAll problem. Machine Learning Pipeline Template case study on a budget. When we  debug the Machine Learning Pipeline on a budget, \ourapproach provides answers with perfect precision, which is also achieved by Explanation Tables with more instances. \ourapproach continues to show higher recall, which increases as the budget increases, and f-measure than Data X-Ray and Explanation Tables.}
  \label{fig:template_budget_all}
\end{figure*}

Similar results hold for the FindAll case as shown in Figure~\ref{fig:template_plot_all} and Figure~\ref{fig:template_budget_all}.

\paragraph{Discussion.} Sometimes, when Data X-Ray uses the instances generated by \ourapproach, it does better, at least in recall. This is expected for the case where the root causes are conjunctions of \modelparameter-comparator-value triples since Data X-Ray was designed to find relevant conjunctions. That is, Data X-Ray  produces a conjunction of \modelparameter-value combinations that lead to bad scenarios. By contrast, \ourapproach finds minimal decision trees for the data seen so far, prioritizing disjunction. When there is little data, jumping to generalizations can be a bad strategy (a lesson we have all learned from real life). This underscores the importance of doing a systematic and iterative search to obtain more data.

The answers provided by Explanation Tables represent a prediction of the pipeline instance evaluation result expressed as a real number. Here, we consider explanations whose prediction is $1.0$, which means a certain, bad result. Therefore, the precision of Explanation Tables is always high, but the recall is usually low. 

\ourapproach dominates the other methods and the performance difference increases as the budget grows. This can be explained by the fact that Data X-Ray provides explanations that are not minimal definitive root causes and Explanation Tables do not handle negations and inequalities.

\section{Related Work} \label{sec:relatedWork}

Recently, the problem of explaining query results and interesting features in data
has received substantial attention in the literature~\cite{Wang:2015:DXD:2723372.2750549, Bailis@sigmod2017, Chirigati:2016:DPM:2882903.2915245,DBLP:journals/pvldb/MeliouRS14, GebalyFGKS14}. 
Some works have focused on explaining where and how errors occur in the data generation process~\cite{Wang:2015:DXD:2723372.2750549} and which data items are more likely to be causes of relational query outputs~\cite{DBLP:journals/pvldb/MeliouRS14,Wang:2017:QDE:3035918.3035925}.  Others have attempted to use data to explain \emph{salient} features in data (e.g., outliers) by discovering relationships between attribute values~\cite{Bailis@sigmod2017, Chirigati:2016:DPM:2882903.2915245, GebalyFGKS14}. 
These approaches have either focused on using data, including provenance, to explain data or considered pipelines consisting of relational algebra operations.
In contrast to the goals of these approaches, \ourapproach aims to diagnose abnormal behavior in
machine learning pipelines that may result from any source of error: data, programs, or sequencing of operations.

Our work is related algorithmically to approaches from hyper-parameter tuning, workflow debugging, and denial constraint identification.
Hyper-parameter tuning methods explore the parameter space of pipelines to optimize their outcome -- they automatically derive instances with improved performance. While their goal is to find good combinations for parameter values, they do not provide any insights into which combinations are most responsible for that performance, which would be analogous to what a high precision debugger finds. 

Prior work on workflow debugging aims to identify and explain problems based on existing provenance, but they do not iteratively derive and test new workflow instances. As we demonstrated in Section~\ref{sec:experiments}, \ourapproach derives good explanations starting the debugging process from scratch and generating fewer pipeline instances than hyper-parameter optimization frameworks. Overall, \ourapproach also gives better recall and precision than non-iterative workflow-debugging tools.

Our approach is also related to the discovery of denial constraints in relational tables (denial constraints are generalizations of common constraints like functional dependencies)~\cite{BleifuB2017}. The main similarity is that both Denial constraints and the diagnoses of \ourapproach  identify conjunctions of parameter-value inequalities.  The Hydra algorithm in particular performs what it calls "focused sampling" to find tuples that satisfy a predicate. \ourapproach does something similar in spirit when looking for pipeline instances that may disprove a hypothesis.

\paragraph{Hyperparameter Optimization.}
Methods based on Bayesian optimization  are considered the state of the art for the hyperparameter optimization problem~\cite{Bergstra2011, Bergstra:2013:MSM:3042817.3042832,Snoek:2012:PBO:2999325.2999464,Snoek:2015:SBO:3045118.3045349,Dolatnia2016}. They can outperform manual setting of parameters, grid search or random search~\cite{Bergstra:2012:RSH:2188385.2188395}. These methods approximate a probability model of the performance outcome given a parameter configuration that is updated from a history of executions. Gaussian Processes and Tree-structured Parzen Estimator are examples of probability models~\cite{Bergstra2011} used to optimize an unknown loss function using the 'expected improvement' criterion as acquisition function. To do this, they assume the search space is smooth and differentiable. This assumption, however, does not hold in general for arbitrary pipelines. Moreover, we are not interested in identifying a bad configurations (we have those to begin with if there have been some bugs already), but in finding minimal root causes.

\paragraph{Debugging and Predicting Pipelines.}
Previous work on pipeline debugging (not limited to machine learning) has focused on analyzing execution history with the goal of identifying problematic parameter settings or inputs. Because they do not use an iterative approach to derive new instances (and associated provenance), they can miss important explanations and also derive incorrect one. That said, the analytical portion of \ourapproach uses many similar ideas to pipeline debugging.

Bala and Chana~\cite{Bala:2015:IFP:2775763.2776365} applied several machine learning algorithms (Na\"{i}ve Bayes, Logistic Regression, Artificial Neural Networks and Random Forests) to predict whether a particular computational pipeline instance will fail to execute in a cloud environment. The goal is to reduce the consumption of expensive resources by recommending against executing the instance if it has a high probability of failure. The system does not try to find the root causes of failure.

The system developed by Chen et al.~\cite{Chen2017} identifies problems by differentiating between  provenance (encoded as trees) of good runs  and bad ones. They then find differences in the trees that may be the reason for the problems. However, the trees do not necessarily provide a succinct explanation for the problems, and there is no assurance that the differences found correspond to root causes.

Viska~\cite{Gudmundsdottir2017} allows users to define a causal relationship between workflow performance and system properties or software versions. It provides big data analytics and data visualization tools to help users to explore their assumptions. Each causality relation defines a treatment (causal variable) and an outcome (performance measurement), the approach is limited to analyze one binary treatment at a time with user in the loop.  

The Molly system~\cite{DBLP:conf/sigmod/AlvaroRH15} combines the analysis of lineage with SAT solvers to find bugs in fault tolerance protocols for distributed systems. Molly simulates failures, such as permanent crash failures, message loss and temporary network partitions, specifically to test fault tolerance protocols over a certain period of (logical clock) time. The process considers all possible combinations of admissible failures up to a user-specified level (e.g., no more than two crash failures and no more message losses after five minutes). While the goal of that system is very specific to fault tolerance protocols, its attempt to provide completeness has influenced our work. In the spirit of  Molly, \ourapproach tries to find \emph{minimal definitive root causes}.

\paragraph{Explaining Pipeline Results.}
Although not designed for pipelines, Data X-Ray~\cite{Wang:2015:DXD:2723372.2750549} provides a mechanism for explaining systematic causes of errors in the data generation process. The system finds common features among corrupt data elements and produce a diagnosis of the problems. If we have provenance of the pipeline instances together with error annotations, Data X-Ray's diagnosis would derive explanations consisting of features that describe the parameter-value pairs responsible for the errors. 

Explanation Tables~\cite{GebalyFGKS14} is another data summary that provides explanations for binary outcomes. Like Data X-Ray, it forms its hypotheses based on given executions, but does not propose new ones. 
Based on a table with some categorical columns (attributes) and one binary column (outcome), the algorithm produces interpretable explanations of the causes of the outcome in terms of the attribute-value pairs combinations.
Explanation Tables express their answers as a disjunction of patterns and each pattern is a conjunction of attribute-equality-value pairs. 

As discussed in Section~\ref{sec:experiments}, \ourapproach produces explanations that are similar to those of Data X-Ray and Explanation Tables, but they are also minimal and are able to handle inequalities and negations. As mentioned above, \ourapproach employs a systematic method to automatically generate new instances that enable it to derive concise explanations that are root causes for a problem.

\section{Conclusion}\label{sec:conclusion}

\ourapproach uses techniques from explanation systems and hyperparameter optimization approaches to address one of the most cumbersome tasks for data scientists and engineers: debugging machine learning pipelines. As far as we know, \ourapproach is the first method that iteratively finds minimal definitive root causes. 

Compared to the state of the art, \ourapproach makes no statistical assumptions (as do Bayesian optimization approaches), but nevertheless achieves higher precision and recall given the same number of pipeline instances. 

There are two main avenues we plan to pursue in future work. First, we would like to make \ourapproach available on a wide variety of systems that support pipeline execution to broaden its applicability. Second, we will use group testing to identify problematic subsets of datasets when a dataset has been identified as a root cause.

\vspace{-.1cm}
\paragraph{Acknowledgments.} We thank the Data X-Ray authors and the Explanation Tables authors for sharing their code with us. This work has been supported in part by the U.S. National Science Foundation under grants MCB-1158273, IOS-1339362, and MCB-1412232, the Brazilian National Council for Scientific and Technological Development (CNPq) under grant 209623/2014-4, the DARPA D3M program the Moore-Sloan Foundation, and NYU WIRELESS. Any opinions, findings, and conclusions or recommendations expressed in this material are those of the authors and do not necessarily reflect the views of funding agencies.

\bibliographystyle{ACM-Reference-Format}
\bibliography{paper}
\end{document}